\documentclass[letterpaper, 10 pt, conference]{ieeeconf}  

\IEEEoverridecommandlockouts                              

\usepackage{amsmath} 
\usepackage{bm}

\usepackage{color}
\usepackage{graphicx}
\usepackage[caption=false,font=normalsize,labelfont=sf,textfont=sf]{subfig}
\usepackage{caption}
\usepackage{stfloats}
\usepackage{float} 
\usepackage{here}
\usepackage{cite}

\usepackage[whole]{bxcjkjatype}

\usepackage{xspace}
\newcommand{\etal}{\textit{et al.}\@\xspace} 

\newcommand{\commentout}[1]{}

\title{\LARGE \bf
Robot Swarm Control Based on Smoothed Particle Hydrodynamics\\for Obstacle-Unaware Navigation*
}

\author{Michikuni Eguchi$^{1,2}$, Mai Nishimura$^{3}$, Shigeo Yoshida$^{3}$, and Takefumi Hiraki$^{1}, \textit{Member, IEEE}$
\thanks{*This work was partially supported by JST ASPIRE Grant Number JPMJAP2327.}
\thanks{$^{1}$Michikuni Eguchi and Takefumi Hiraki are with Cluster Metaverse Lab,
        8-9-5 Nishigotanda, Shinagawa, Tokyo 140-0031 Japan
        {\tt\small \{m.eguchi, t.hiraki\}@cluster.mu}}%
\thanks{$^{2}$Michikuni Eguchi is with the Graduate School of Comprehensive Human Sciences, University of Tsukuba,
        1-2 Kasuga, Tsukuba, Ibaraki 305-8550 Japan}%
\thanks{$^{3}$Mai Nishimura and Shigeo Yoshida are with OMRON SINIC X Corporation,
        5-24-5 Hongo, Bunkyo, Tokyo, 113-0033 Japan
        {\tt\small \{mai.nishimura, shigeo.yoshida\}@sinicx.com}}%
}

\begin{document}

\maketitle
\thispagestyle{empty}
\pagestyle{empty}

\begin{abstract}
Robot swarms hold immense potential for performing complex tasks far beyond the capabilities of individual robots.
However, the challenge in unleashing this potential is the robots' limited sensory capabilities, which hinder their ability to detect and adapt to unknown obstacles in real-time.
To overcome this limitation, we introduce a novel robot swarm control method with an indirect obstacle detector using a smoothed particle hydrodynamics (SPH) model.
The indirect obstacle detector can predict the collision with an obstacle and its collision point solely from the robot's velocity information.
This approach enables the swarm to effectively and accurately navigate environments without the need for explicit obstacle detection, significantly enhancing their operational robustness and efficiency.
Our method's superiority is quantitatively validated through a comparative analysis, showcasing its significant navigation and pattern formation improvements under obstacle-unaware conditions. 
\end{abstract}

\section{INTRODUCTION}
The use of large groups of robots, as opposed to individual robots, to perform complex tasks has attracted considerable interest.
These groups, referred to as \emph{robot swarms}, draw inspiration from natural phenomena where animal collectives interact to coordinate behavior and demonstrate collective intelligence that enables them to accomplish tasks beyond the capabilities of individual members~\cite{Sahin2005-ik, Brambilla2013-bg}.
Distinctive aspects of robot swarms are their ability to effectively operate in environments with unknown obstacles, their scalability, and their reliance on autonomous decentralized control based on local interactions rather than on centralized control.

Owing to their small size, which inherently limits their sensory capabilities, most robots are only able to collect limited environmental data. 
Although swarm robotics holds immense potential in applications such as teleoperation interfaces for human-computer interaction (HCI)~\cite{Suzuki2022-xv}, it still requires significant departures from staging environments.
In real-world applications, the robots lack any prior knowledge about the environment and often encounter unexpected obstacles that they cannot perceive owing to their limited sensory capabilities.
This significantly limits their ability to adapt to unknown environments, leading us to a key question: ``\emph{Can we navigate a swarm of robots without the need for explicit obstacle localization?''}

We introduce a new concept named \emph{obstacle-unaware navigation}.
Given only the positions and velocities of the robots, with no sensory perception of any obstacles, our aim is to efficiently navigate a robot swarm toward their destination while avoiding obstructions in the environment.
To this end, this study proposes a new control method enabling robot swarms to navigate through and form patterns in complex environments by utilizing indirect obstacle detection.
Our proposed method allows robot swarms to avoid obstacles and organize themselves into specific patterns and shapes in complex environments, especially when obstacles cannot be directly detected.

\begin{figure}[t]
    \centering
    \includegraphics[width=\columnwidth]{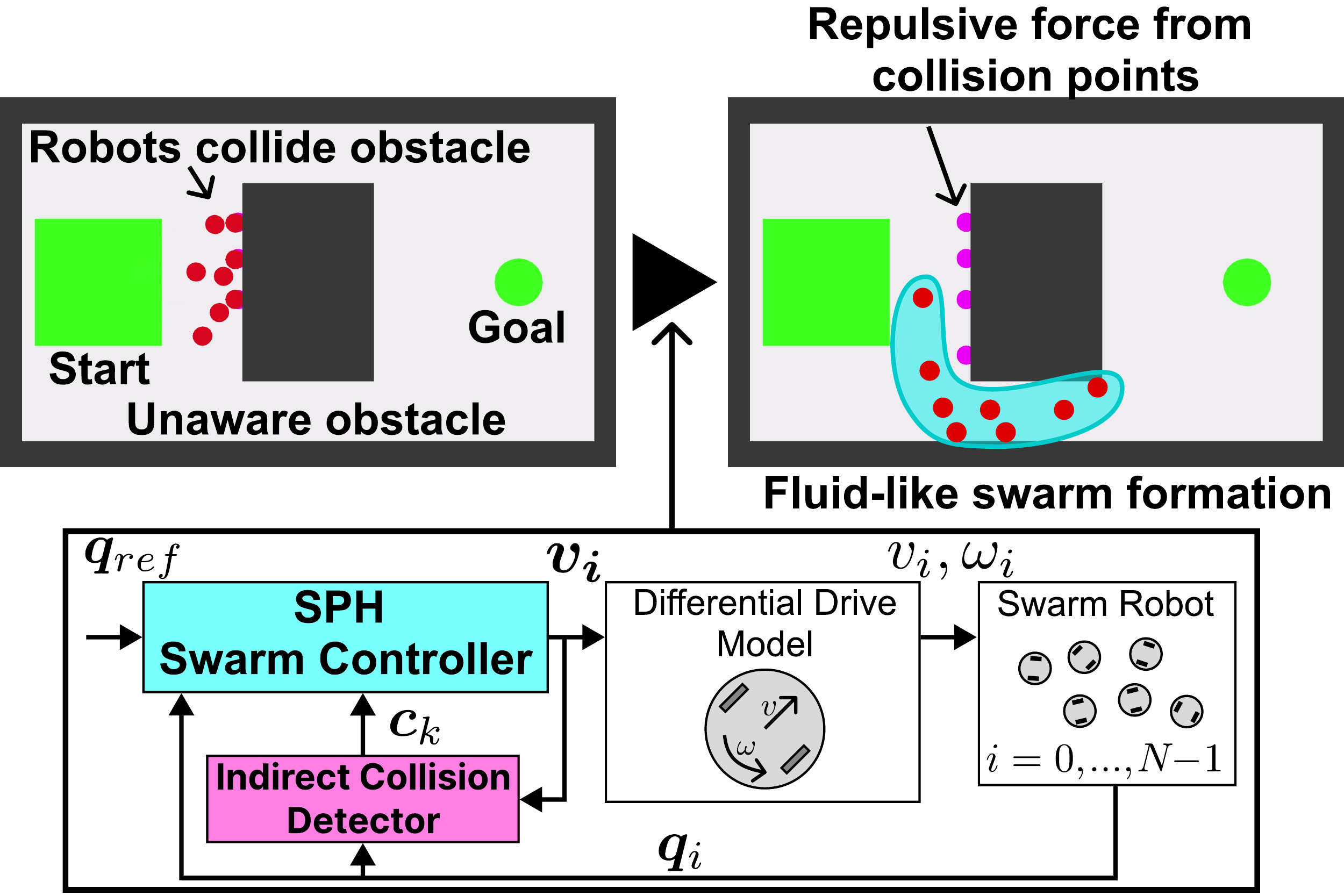}
    \caption{Conceptual image of the situation we are challenged to solve and our proposed method. When robots collide with an undetected obstacle, they detect the collision indirectly from the change in their velocity. We add an element of repulsion force from these collision points to the SPH-based controller to achieve obstacle-unaware navigation.}
    \label{fig:controller_outline}
\end{figure}

Under the concept of indirect obstacle detection, the robots themselves detect that they are in a collision state based on the difference between the commanded and observed velocity values, without using external sensors.
We integrate this collision detector into a feedback control method based on the smoothed particle hydrodynamics (SPH) model, which conceptualizes the robot swarm as a fluid.
In particular, by integrating the repulsive force when the robots encounter obstacles into this SPH model, we equip the robots with the ability to adapt to obstacle-unaware environments.
Fig.~\ref{fig:controller_outline} shows the situation we intend to solve and the proposed control method.
This advancement enhances the capability of the swarm to robustly operate in environments where obstacles cannot be directly detected, thereby significantly increasing the swarm's usefulness in various applications, including HCI.

We also demonstrate the effectiveness of our approach through a comparative analysis with conventional navigation methods in obstacle-unaware scenarios.
The results from both the simulations and real robot experiments confirm that our proposed method significantly improves the applicability of swarm robotics and provides a solid framework for navigation and patterning in obstacle-unaware environments.

\section{RELATED WORK}
\subsection{Collision Avoidance}\label{sec:collision_avoidance}
Collision avoidance has received significant research attention in robotics owing to its importance in real-world applications.
Many studies have reported the potential function method~\cite{Khatib1986-si} and the graph search method~\cite{Hart1968-ir} owing to the clarity of their theories.
In recent years, collision avoidance methods that use machine learning have alse been reported~\cite{Xie2017-zp}.

Several collision avoidance measures have been reported for robot swarms.
The Reciprocal Velocity Obstacles (RVO) method~\cite{Van_den_Berg2008-we} calculates collision-free velocity commands by considering the velocities of the surrounding robots and obstacles, enabling individual robots to predict and avoid collisions.
Furthermore, the performance of swarm robot control based on deep reinforcement learning (DRL)~\cite{Sui2021-zl} reportedly surpasses that of the RVO method.
This is attributable to DRL's ability to autonomously acquire optimal obstacle avoidance strategies based on environmental information. 

However, note that these collision avoidance methods rely on the accurate acquisition of obstacle location, which may not always be possible.
Therefore, it is difficult for existing collision avoidance methods to satisfactorily demonstrate their performance in situations common to small swarm robots, such as when no sensor is available for obstacle detection.

\subsection{Robot Swarm Control Based on Hydrodynamic Models}
Motion planning and control for robot swarms originated from mimicking biological behaviors observed in insects and birds.
Methods grounded in physical principles emulating the behavior of gases, fluids, and solids have recently been actively explored~\cite{Spears2004-kw}.

Among these methods, the SPH model~\cite{Gingold1977-po}, used in fluid dynamic simulations, is of interest for controlling robot swarms to mimic fluid behavior.
This approach has been effective in applications such as collision avoidance~\cite{Chua2021-bn, Perkinson2005-px}, trajectory tracking~\cite{Lipinski2011-ie}, and swarm shape control~\cite{Zhao2011-nn}.
In particular, Pac~\etal~\cite{Pac2007-op} and Pimenta~\etal~\cite{Pimenta2013-mz} have proposed the use of the SPH method to model a robot swarm as a fluid analogy, allowing for the control of swarm flow by adjusting the flow parameters.
These methods enable control over the behavior of the robot swarm via the parameters in the governing equations.
Our method builds upon this notion, extending the capabilities of SPH-based methods.

The SPH model shares several benefits with other physical concept-based controls for swarm robots.
However, its advantage is its decentralized and scalable approach, which requires only information from robots in the local neighborhood.
Although some methods~\cite{Arkin1997-ue, Howard2002-vr} offer this advantage, the SPH method stands out for its unique feature to control the density of robots.
This feature makes the method suitable for controlling swarms to a specific target position or region.

In SPH-based robot swarm control, real-time obstacle avoidance can be achieved utilizing high-density particles at the boundaries of detected obstacles~\cite{Pac2007-op}.
However, this assumes that obstacles are detected in advance, which poses challenges in situations where obstacles are indirectly detected when robots collide and subsequently get stuck.

\subsection{Robot Swarm Application Using Obstacle-Unaware Navigation}
Real-world applications leveraging small robot swarms' robustness, flexibility, and scalability have been extensively explored~\cite{Rubenstein2014-dm, Alonso-Mora2011-hs}.
Notably, the Swarm User Interface (Swarm UI) concept, first introduced through Zooids~\cite{Le_Goc2016-fd}, has focused on expanding robot functionalities and integrating mixed-reality systems for enhanced interaction~\cite{Hiraki2018-dj, Ichihashi2024-gs}.
Moreover, small flying robot swarms have shown promise in display technologies via formation flying~\cite{Ghandeharizadeh2022-dv}.

A challenge for these applications is the limited sensory capability of the swarm owing to the robot's size, necessitating control mechanisms that do not rely on detailed environmental data, such as obstacle or collision information.
Scalability and autonomous operation are achieved through decentralized control predicated based on local interactions.
For navigation and obstacle avoidance, swarms often employ potential field-based control methods~\cite{Arkin1989-uy}, with several Swarm UI implementations adopting projective control strategies for enhanced spatial manipulation~\cite{Le_Goc2016-fd, Hiraki2018-dj, Ichihashi2024-gs}.

Although these control methods are effective in obstacle-free environments, real-world conditions present numerous unforeseen obstacles, leading to potential control challenges.
Some approaches have allowed intentional collisions with obstacles, especially in small flying robot swarms~\cite{Briod2013-sx, Mulgaonkar2018-eg}.
However, these studies have focused on collision avoidance and have not evolved to robot swarm control that can adaptively update the controller with the information after detecting an obstacle owing to collision to reach a goal or form a desired shape or pattern.

\section{METHOD}
\subsection{Controller based on Smoothed Particle Hydrodynamics}
\label{sec:SPH}
In this study, we extend a controller for swarm robot navigation using the SPH approach, initially introduced in~\cite{Pac2007-op}.
This subsection elaborates on the SPH-based approach for controlling robot swarms.
In the SPH approach, fluids are modeled through a multitude of discrete particles, where each particle embodies the fluid's physical characteristics and contributes to simulating fluid dynamics by tracking changes in these characteristics over time.
The SPH-based controller for robot swarms replaces particles with robots to achieve swarm robot control such that the robot swarm behaves like a fluid.
The SPH method is known for its ability to approximate the physical quantities of the fluid at any given point by aggregating the attributes of the neighboring robots, using a kernel function.
This function weighs the contributions of robots based on their proximity, facilitating the smooth estimation of local physical phenomena.
We adopt the Gaussian kernel, denoted by $W(R_{ij})$, for this purpose, can be expressed as follows:
%
\begin{equation} \label{eq:sph_kernel_function}
    \begin{aligned}
    W(R_{ij}) &= 
    \begin{cases} 
        \alpha_d e^{-\frac{R_{ij}^2}{h^2}} & \text{for } R_{ij} \leq \kappa h \\
        0 & \text{otherwise}
    \end{cases} \\
    \alpha_d &= \frac{1}{\pi h^2}, \quad
    R_{ij} = \frac{\| \bm{q}_i - \bm{q}_j \|}{h}, \quad
    \kappa = 2
    \end{aligned}
\end{equation}


where $R_{ij}$ represents the normalized distance between the 
two-dimensional (2D) position $\bm{q}_i=(x_i,y_i)$ of a robot $i$ and that of another robot $j$ within a swarm, comprising of $N$ robots in total.
The parameter $h$, known as the smoothing length, defines the spatial extent influenced by the kernel.
In the SPH model, each robot's density $\rho_i$ is calculated as shown in (\ref{eq:density}).
\begin{equation} \label{eq:density}
\rho_i = \sum_{j=0}^{N-1} m_j W(R_{ij}),
\end{equation}
Here, $\rho_i$ represents the density at robot $i$, calculated by summing the products of the masses $m_j$ of all robots $j$ and the kernel function $W(R_{ij})$, which adjusts the influence based on distance.
The fluid force $\bm{f}^{\mathrm{sph}}_{i}$ acting on each robot in the swarm can be expressed as follows.
\begin{equation} \label{eq:sph_force}
    \bm{f}^{\mathrm{sph}}_i = \sum_{j=0}^{N-1} m_j \left( \frac{\mathbf{\sigma}_i}{\rho_i^2} + \frac{\mathbf{\sigma}_j}{\rho_j^2} \right) \nabla_i W_{ij}
\end{equation}
Here, $\nabla_i W_{ij} \in \Re^2$ represents the gradient of the kernel function with respect to robot positions, and $\sigma_i$ denotes the stress tensor, incorporating both pressure and viscous contributions as described in the following equations.
\begin{equation} \label{eq:stress}
\begin{aligned}
    \sigma^{xx}_i =& -p_i + \mu_i \left( 2 \frac{\partial v_{i}^x}{\partial x} - \frac{2}{3} \left( \frac{\partial v_{i}^x}{\partial x} + \frac{\partial v_{i}^y}{\partial y} \right) \right)
    \\
    &\sigma^{xy}_i = \mu_i \left( \frac{\partial v_{i}^y}{\partial x} + \frac{\partial v_{i}^x}{\partial y} \right)
    \\
    \sigma^{yy}_i =& -p_i + \mu_i \left( 2 \frac{\partial v_{i}^y}{\partial y} - \frac{2}{3} \left( \frac{\partial v_{i}^x}{\partial x} + \frac{\partial v_{i}^y}{\partial y} \right) \right),
\end{aligned}    
\end{equation}
where the total stress tensor $\sigma_i$ is expressed in terms of fluid pressure and viscous stress.
The pressure $p_i$ produces repulsive and attractive forces between robots and creates a flow that directs the swarm toward uniform distribution.
Viscous stress is expressed as the product of the viscosity coefficient $\mu_i$ and the velocity of the robots $\bm{v}_{i}=(v_i^x,v_i^y)$, and generates an attractive force that makes the velocity vector of the entire swarm uniform.

Because this study uses the model of incompressible fluids such as water~\cite{Zhao2011-nn}, the pressure $p_i$ is expressed as follows:
\begin{equation}
\label{eq:pressure_incompressible}
     p_i = K\rho_0 \left[ \left(\frac{\rho_i}{\rho_0}\right)^\gamma - 1 \right],
\end{equation}
where $K$ represents the stiffness constant, $\rho_0$ is the reference density, and $\gamma$ is the adiabatic constant, typically set to 7 to match the convenient incompressibility as suggested by Zhao~\etal~\cite{Zhao2011-nn}.

We incorporate two additional forces to facilitate collision avoidance within the swarm and position control for navigating the swarm to the target point.
The repulsive force $\bm{f}_{i}^\mathrm{rep}$ between the robots to prevent robot collisions can be expressed as follows:
\begin{equation} \label{eq:force_repulsive}
    \bm{f}_{i}^{\mathrm{rep}} = K_\mathrm{rep} \sum_{j = 0}^{N-1}  \frac{\bm{q}_{i}-\bm{q}_{j}}{\| \bm{q}_{i}-\bm{q}_{j} \|^2} W(R_{ij}),
\end{equation}
where $K_{\mathrm{rep}}$ is the repulsive gain between the robots.
The position control force $\bm{f}_{i}^{\mathrm{pos}}$ is introduced to control the position of the swarm to reach the target point.
Here, position control is based on the Proportional-Differential (PD) control law for the error $\bm{e}_i$ between the target point $\bm{q}_{\mathrm{ref}}$ and $\bm{q}_{i}$, which is expressed as follows:
\begin{equation} \label{eq:force_position}
    \begin{aligned}
        \bm{f}_{i}^{\mathrm{pos}} =& K_p  \bm{e}_i - K_d \frac{d\bm{e}_i}{dt}
        \\
        \bm{e}_i &= \bm{q}_{\mathrm{ref}} - \bm{q}_{i},
    \end{aligned}
\end{equation}
where $K_p$ and $K_d$ are the proportional and derivative gains for PD control, respectively.

From (\ref{eq:sph_force}), (\ref{eq:force_repulsive}), and (\ref{eq:force_position}), the total force applied to each robot, which combines the fluid dynamic forces with repulsive and position control forces, is calculated as follows:
\begin{equation} \label{total_force_prev}
    \frac{d\bm{v}_{i}}{dt} = \bm{f}^\mathrm{sph}_{i} + \bm{f}_{i}^\mathrm{rep} + \bm{f}_{i}^\mathrm{pos}.
\end{equation}

Finally, the velocity vector $\bm{v}_{i}$, which is the input to each robot, is updated by the forward Euler method at sampling time $\Delta t$ and can be expressed as follows:
\begin{equation} \label{eq:forward_euler}
    \bm{v}_{i}(t+\Delta t) = \bm{v}_{i}(t) +  \frac{d\bm{v}_{i}(t)}{dt} \Delta t
\end{equation}
By integrating the total force from (\ref{total_force_prev}) into this update equation, we achieve an SPH-based control mechanism that simulates the behavior of an incompressible fluid, enabling effective swarm control.

\subsection{SPH-based Controller with Indirect Obstacle Detection}
We extend the SPH-based controller to explore robot swarm control without collision avoidance using indirect obstacle detection.
In particular, we develop a robot swarm controller that achieves obstacle avoidance by leveraging the difference between the commanded and observed velocity that occurs when robots collide with an obstacle and get stuck, and by adding a repulsion force from the point of contact with the obstacle to the SPH model.

\subsubsection{Indirect Collision Detection}
Conventionally, collision detection has been mainly performed by contact sensors installed on the robot's exterior~\cite{Petris2021-qv} or by acceleration judgments during collisions~\cite{Briod2013-sx}.
However, adding sensors for external sensing in small robot swarms such as those in this study owing to size constraints.
Also, in the case of acceleration judgments, for low-speed robots such as those in this study, the difference in acceleration between normal running and collision is not significant, resulting in false collision detection.

Therefore, we propose a new collision detection method using the time integral of the tracking error against the robot's command velocity.
When the robot runs normally, its observed velocity $\bm{v'}$ follows the command velocity $\bm{v}$.
\begin{equation} \label{eq:observed_velocity}
    \bm{v'} = \frac{\bm{q}(t) - \bm{q}(t-\Delta t)}{\Delta t} 
\end{equation}

On the other hand, when the robot collides with an obstacle, its position is constrained by the obstacle, and the observed velocity is close to zero, resulting in a large difference between the command velocity and the observed velocity.
In this method, the difference between the commanded and observed velocities is added (integrated) at each control time.
The integral value $I(t+\Delta t)$ at time $t+\Delta t$ can be expressed as follows:
\begin{equation} \label{eq:collision_detection_model}
    \begin{aligned}
        I(t+\Delta t) = I(t) + \frac{\left|\|\bm{v}\| - \|\bm{v'}\| \right|}{V_{max}} - \zeta \\
        I(t+\Delta t)\leftarrow0 \quad \text{if } I(t+\Delta t) \geq I_{thr}.
    \end{aligned}
\end{equation}

If this integral value exceeds the threshold value $I_{thr}$, that is, $I(t+\Delta t) \geq I_{thr}$, a collision is detected, and the robot's position $\bm{q}$ at that time is recorded as the new collision position $\bm{c}$.
The maximum velocity $V_{max}$ of the robot is utilized to normalize velocity values.
The attenuation value $\zeta$ is a constant value subtracted to suppress the effect of the tracking error that occurs during the normal running of the robot on the velocity command value.
For example, even a robot with poor tracking performance can detect a collision with high accuracy by increasing $\zeta$.

\subsubsection{SPH with Obstacle Avoidance}
To realize SPH-based robot swarm control with obstacle avoidance, the SPH-based controller is modified to generate repulsion force from the point of contact with the obstacle.
Let $\bm{c}_k$ be the collision point $k$, the repulsion force $\bm{f}_{i}^{\mathrm{obs}}$ acting on robot $i$ from $M$ collision points can be expressed as follows:
\begin{equation} \label{eq:force_obstacle}
    \begin{aligned}
        \bm{f}_{i}^{\mathrm{obs}} = K_{\mathrm{obs}} &\sum_{k = 0}^{M-1}  \frac{\bm{q}_{i}-\bm{c}_{k}}{\| \bm{q}_{i}-\bm{c}_{k} \|^2} W(R_{ik})\\
        R_{ik} &= \frac{\| \bm{q}_i - \bm{c}_k \|}{h},
    \end{aligned}
\end{equation}
Here, $K_{\mathrm{obs}}$ is the gain of the repulsion at the collision point.
By combining this repulsion with the forces introduced by the SPH described in Section~\ref{sec:SPH}, the virtual force on each robot in the swarm can be represented as follows:
\begin{equation} \label{total_force}
    \frac{d\bm{v}_{i}}{dt} = \bm{f}^{\mathrm{sph}}_{i} + \bm{f}_{i}^{\mathrm{rep}} + \bm{f}_{i}^{\mathrm{pos}} + \bm{f}_{i}^{\mathrm{obs}}.
\end{equation}
By controlling a robot swarm with an SPH-based controller modified to avoid this obstacle, robot swarm control can be achieved even in environments where the obstacle is unknown.

\subsection{Kinematics of a Differential Drive Robot}
Most of the robot swarms we consider in this study comprise two-wheeled differential drive models owing to their size constraints.
Therefore, we must consider the nonholonomic constraints of the differential drive model.
%
To address this nonholonomic constraint, we adopt the effective center model, an approach proposed by Snape~\etal~\cite{Snape2010-sx}.
In this model, the origin of the robot's coordinate system is a point offset from the robot's center by $d$ in the direction of the robot forward.
The robot is considered a holonomic robot with an effective center at that point and a radius $R = r + d$, where $r$ is the radius of the robot. This holonomic robot is controlled by velocity vector $\bm{v}$.
The conversion from the velocity vector $\bm{v}$ to the input $v,\omega$ of the differential drive model is expressed by the following equation:
\begin{equation} \label{eq:feedback_linearization}
\begin{bmatrix}
v \\
\omega
\end{bmatrix}
=
\begin{bmatrix}
\cos(\theta) & \sin(\theta) \\
-\frac{\sin(\theta)}{d} & \frac{\cos(\theta)}{d}
\end{bmatrix}
\begin{bmatrix}
v^x \\
v^y
\end{bmatrix}
.
\end{equation}
where $\theta$ is the orientation of the robots.
We calculate $v,\omega$ using (\ref{eq:feedback_linearization}) and use these values to control the motors of the robot swarm. 


\section{EXPERIMENTS}
The evaluation experiment aimed to determine whether the proposed swarm robot controller can navigate environments with obstacles without obstacle information and how efficiently it can handle complex environments compared with conventional methods.
Therefore, we conducted experiments in both simulation and real settings to verify the probability of and the time required for reaching the target in each environment when the robot swarm moved in multiple environments.

\subsection{Baseline Control Methods}
\begin{figure}[t]
    \centering
    \includegraphics[width=\columnwidth]{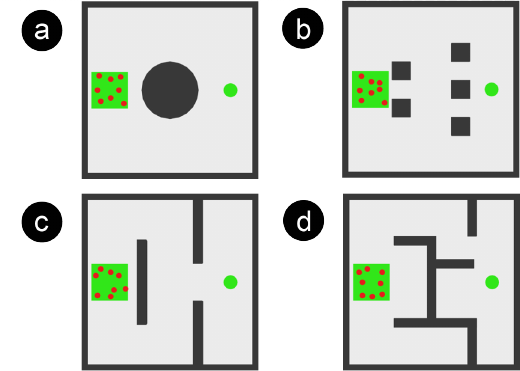}
    \caption{Appearance of the four simulation field environments: (a) Entry, (b) Dense pillar, (c) Barricade, (d) Pocket maze. The robot swarm (red dots) is initially located within the lime green square on the left side of the field. The robots run to the goal point, shown as a lime green dot on the right side, avoiding obstacles shown in black.}
    \label{fig:simulation_setup}
\end{figure}
We used the normal \textbf{SPH}-based method~\cite{Pac2007-op}, the \textbf{RVO}~\cite{Van_den_Berg2008-we} method, and the contact-based obstacle avoidance method~\cite{Mulgaonkar2018-eg} (hereafter referred to as \textbf{Bound}) as our baseline robot swarm control method.
The normal SPH-based method is described in Section~\ref{sec:SPH}.

RVO is the de facto standard method in swarm robot control as mentioned in Section~\ref{sec:collision_avoidance}, and it has been adopted in many swarm robot interfaces ~\cite{Le_Goc2016-fd, Ichihashi2024-gs}.
To verify how RVO-based swarm control behaves in an obstacle-unaware environment, we provided the RVO method with only the information about each robot and its target position and no information about the obstacles in the environment.

Bound is a collision-aware control method that allows robots to avoid obstacles by moving to repel from the point of collision with the obstacle.
In swarm control using this method, the interaction between robots is not specifically considered, so each robot is individually controlled.
We employed this method to confirm the obstacle avoidance performance of the method using only collisions without considering the interaction of the swarm.

\subsection{Experimental Setup}
\subsubsection{Field Environments}
\begin{figure*}[t]
    \centering
    \includegraphics[width=\linewidth]{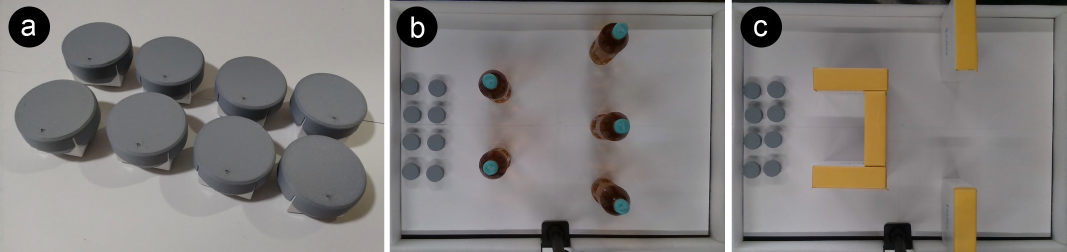}
    \caption{Appearance of the robots and the experimental environment; (a) the eight swarm robots used in the experiment, (b, c) the environment imitating (b) the Dense pillar, and (c) the Pocket maze in the simulation.}
    \label{fig:experiment_setup}
\end{figure*}
We designed four environments for the simulation experiments (Fig.~\ref{fig:simulation_setup}) and two environments for the real experiments (Fig.~\ref{fig:experiment_setup}b, c).
The dimensions of each simulation and real environment were 0.9 m $\times$ 0.9 m and 0.6 m $\times$ 0.8 m, respectively.
Each environment was constructed as a field containing obstacles in different configurations to verify obstacle avoidance in an obstacle-unaware environment.
The characteristics of each environment are described below.
\begin{itemize}
   \item \textbf{Entry}: The simplest obstacle environment. To compare the arrival time of the swarm with each method, the shape of the obstacle is curved so that the robot does not stop at the obstacle.
   \item \textbf{Dense pillar}: Obstacle environment as described in Bound~\cite{Mulgaonkar2018-eg}.
   \item \textbf{Barricade}: The obstacles are placed in a straight line from the start to the goal point, which is heavily obstructed. Unlike the Entry and Dense pillar environments, the robots here must significantly change their path to avoid the obstacles.
   \item \textbf{Pocket maze}: Pocket-shaped obstacles are placed. The robots must retrace its path to avoid these obstacles.
\end{itemize}

\subsubsection{Robot Swarm and Control System}
\begin{table*}[!b]
    \centering
    \caption{Results of the simulation experiment. We measured the goal reachability rate and the mean time to reach the goal for each environment. Our proposed method showed the best performance in all environments.}
    \label{tab:simulation_result}
        \scalebox{0.9}{
            \begin{tabular}{c|c c|c c|c c|c c}
                &
                \multicolumn{2}{|c|}{\textbf{Entry}} &
                \multicolumn{2}{|c|}{\textbf{Dense pillar}} &
                \multicolumn{2}{|c|}{\textbf{Barricade}} &
                \multicolumn{2}{|c}{\textbf{Pocket maze}} 
                \\
                \textbf{Controllers} &
                \textbf{Reachability rate} &
                \textbf{Mean time} &
                \textbf{Reachability rate} &
                \textbf{Mean time} &
                \textbf{Reachability rate} &
                \textbf{Mean  time} &
                \textbf{Reachability rate} &
                \textbf{Mean time} 
                \\
                \hline
                SPH~\cite{Pac2007-op} & 76 \% (38/50) & 17.64 s & 8 \% (4/50) & 14.80 s & 0 \% (0/50) & -- & 0 \% (0/50) & -- \\
                RVO~\cite{Van_den_Berg2008-we} & 78 \% (39/50) & 16.74 s & 6 \% (3/50) & 14.69 s & 0 \% (0/50) & -- & 0 \% (0/50) & -- \\
                Bound~\cite{Mulgaonkar2018-eg}& 100 \% (50/50) & 22.73 s & 72 \% (36/50) & 10.87 s & 74 \% (37/50) & 41.34 s & 0 \% (0/50) & -- \\
                \textbf{Ours} & \textbf{100 \% (50/50)} & \textbf{5.73 s} & \textbf{98 \% (49/50)} & \textbf{9.69 s} & \textbf{98 \% (49/50)} & \textbf{8.52 s} & \textbf{92 \% (46/50)} & \textbf{15.41 s} \\
            \end{tabular}
        }
\end{table*}

We have verified the performance of the proposed controller by forming a robot swarm with eight small robots (toio, Sony Interactive Entertainment), as shown in Fig.~\ref{fig:experiment_setup}a.
The robots are cube-shaped with a length and width of 32 mm and a height of 25 mm, and they can move on a flat surface using two drive wheels.
We attached a circular cap with a diameter of 45 mm to the top of each robot to reduce the effect of snagging on obstacles.
The robot's control system was programmed in Unity, written in C\#, and executed on a PC (ThinkPad X13 Gen 2, Lenovo, CPU: AMD Ryzen 5 PRO 4650U, RAM: 16GB).
We performed the simulation experiments under the Unity software environment using this PC.
Each robot is also equipped with an optical sensor on its bottom, allowing it to determine its absolute position by operating on a paper mat with a printed dot pattern for position tracking.
All individual robots are controlled by a control system on a PC via Bluetooth communication in 100 ms cycles, for exchanging position and velocity control information.

\subsubsection{Procedure}
We measured the success rate of the robot swarm reaching the goal point and the mean time to reach the goal point in both simulation and real environments, when the robot swarm traveled from the start to the goal point at a maximum speed of 0.2 m/s.
We judged a robot swarm to have reached the goal when all robots in the swarm were within a 0.15 m radius of the goal, and the speed was 0.1 m/s or less.
In all experiments, we set a time limit of 100 seconds from the start of the run and considered a trial a failure if it did not meet the arrival condition within this time limit.
The mean arrival time was calculated using only those trials where the arrival conditions were met.
We performed these experiments 50 times each in the simulation environment and five times each in the real environment for each robot swarm using a different controller.

\subsection{Results}
\subsubsection{Experiment in Simulation Environment}

\begin{figure*}[t]
    \includegraphics[width=\linewidth]{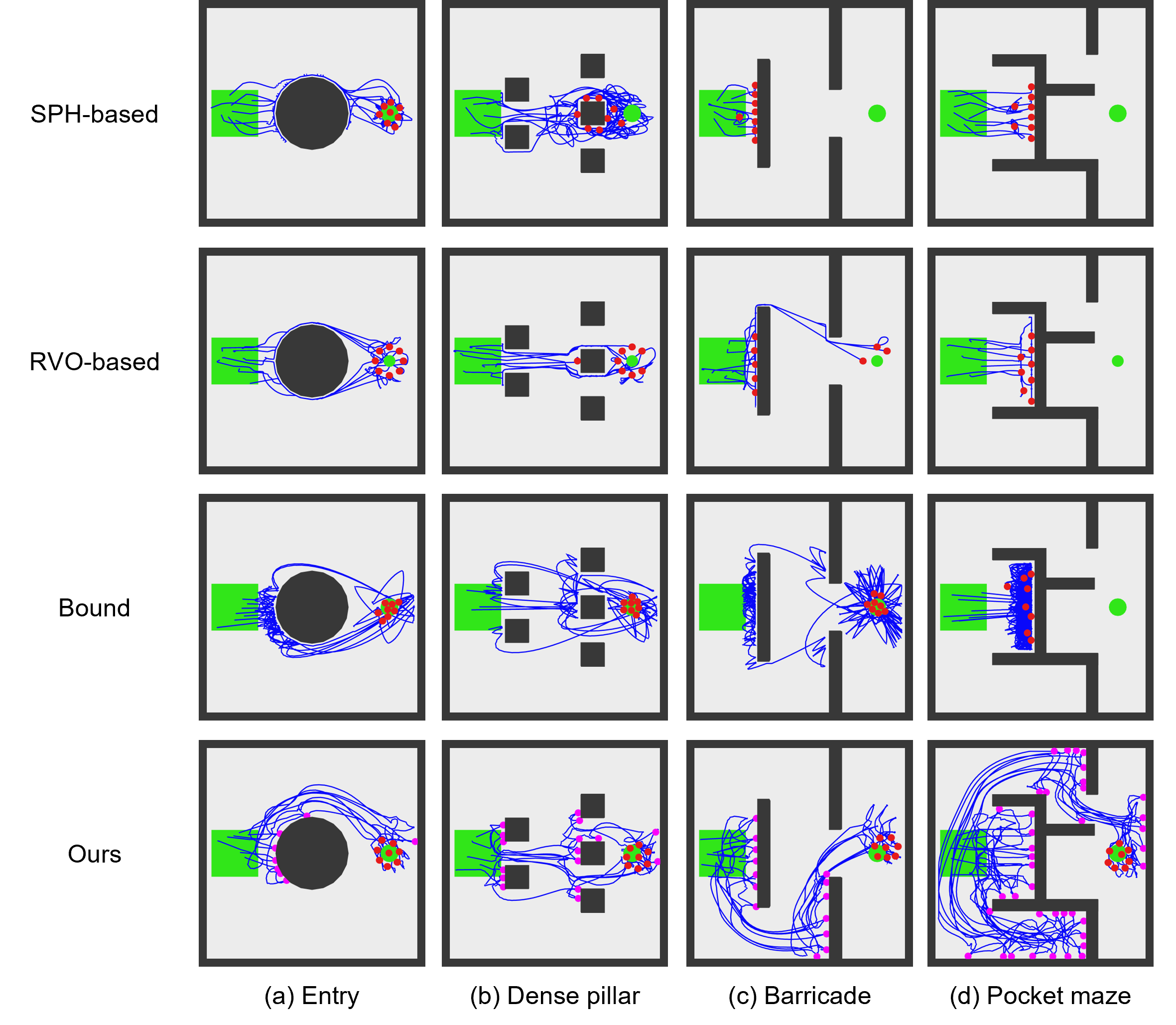}
    \caption{Trajectories of robot swarms in the simulation experiment. The blue lines in the figures represent the running trajectory of each robot, and the red dots represent the positions of the robots at the end of the simulation. The magenta dots represent the detected collision positions using our proposed method.}
    \label{fig:simulation_results}
\end{figure*}

In the simulation experiment, the goal reachability rate and the mean time to reach the goal for each trial of the robot swarm guided by each control method are shown in Table~\ref{tab:simulation_result}.
In addition, the robot swarm's paths across different environments are depicted in Fig.~\ref{fig:simulation_results}.
The proposed method consistently outperformed all baselines, achieving the highest goal reachability rates and the shortest mean times to reach the goal point across all tested scenarios.

Our method achieved over 98\% success in the Entry, Dense Pillar, and Barricade environments, and over 90\% in the complex Pocket maze environment.
This performance underscores our method's superior obstacle navigation capabilities, even in obstacle-unaware environments.
Compared to the baseline methods, which maintained a success rate of above 70\% in the Entry environment, our method's adaptability to more challenging environments such as the Dense Pillar and Barricade was notably superior.

We found that all baseline methods had a success rate of over 70\% in the Entry environment, whereas the Bound method also achieved this success rate in the Dense Pillar and Barricade environments.
In particular, the Bound method, designed for contact-based obstacle avoidance, outperformed others such as RVO and standard SPH in certain environments.
However, its effectiveness waned in complex scenarios requiring navigational retreats, such as the Pocket Maze, where only our proposed method effectively circumvented pocket-shaped obstacles by reversing direction.

In terms of mean arrival time, the proposed method reduced the time to reach the goal in all environments, achieving goals in under one-third the time taken by the baseline methods in the Entry and Barricade environments. 
This efficiency is attributed to our method's proactive avoidance strategy, which prevents prolonged obstacle encounters by steering clear from previous collision points, as illustrated in Fig.~\ref{fig:simulation_results}.

\subsubsection{Experiment in Real Environment}
\begin{table}[b]
    \centering
    \caption{Results of the experiment in real environment. The measured metrics are the same as in the simulation experiment. Our proposed method also showed the best performance.}
    \label{tab:real_result}
    \scalebox{0.82}{
    \begin{tabular}{c|c c|c c}
        &
        \multicolumn{2}{|c|}{\textbf{Dense pillar}} &
        \multicolumn{2}{|c}{\textbf{Pocket maze}} 
        \\
        \textbf{Controllers} &
        \textbf{Reachability rate} &
        \textbf{Mean time} &
        \textbf{Reachability rate} &
        \textbf{Mean time}
        \\
        \hline
        SPH~\cite{Pac2007-op} & 0 \% (0/5) & -- & 0 \% (0/5) & -- \\
        RVO~\cite{Van_den_Berg2008-we} & 0 \% (0/5) & -- & 0 \% (0/5) & -- \\
        Bound~\cite{Mulgaonkar2018-eg}& 100 \% (5/5) & 45.11 s & 0 \% (0/5) & -- \\
        \textbf{Ours} & \textbf{100 \% (5/5)} & \textbf{41.52 s} & \textbf{80 \% (4/5)} & \textbf{68.82 s} \\
    \end{tabular}
    }
\end{table}
\begin{figure}[b]
    \centering
    \includegraphics[width=\columnwidth]{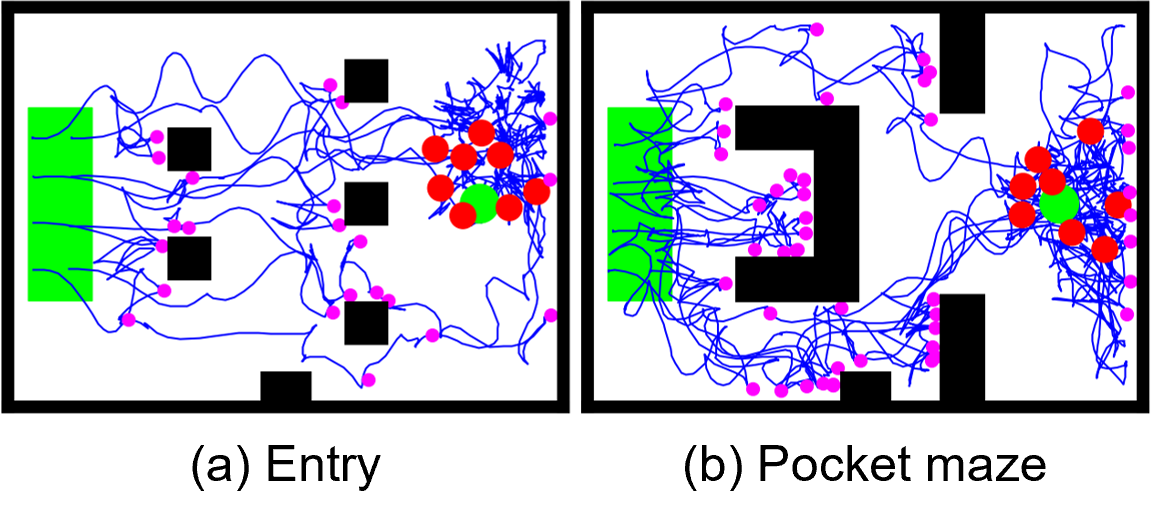}
    \caption{Trajectories of the robot swarm with the proposed controller in the real environments. The content represented by colored lines and dots is the same as in Fig.~\ref{fig:simulation_results}.}
    \label{fig:real_result}
\end{figure}

Table~\ref{tab:real_result} shows the goal reachability and the mean time of the robot swarm to reach the goal for each trial in the real experiment, and the trajectory of the robot swarm with the proposed controller in each environment is shown in Fig.~\ref{fig:real_result}.
In the real experiment, as in the simulation environment, the proposed method reached the goal with the highest reachability rate and the shortest mean arrival time.

However, some differences were noted between the actual and simulation results.
First, the collision points detected via indirect obstacle detection were not only near the obstacles but also near other obstacles.
This is probably attributable to the velocity control performance of the actual machine being worse than that of the simulator, which could be overcome by increasing the detection threshold $I_{th}$ and the attenuation value $\zeta$ of the indirect obstacle detection.
However, this change would increase the time to collision detection, creating a trade-off between detection accuracy and time.
In addition, the position convergence of the robots was low near the target arrival position, resulting in vibrations.
This is thought to be attributable to a large lag in the actual velocity of the two-wheel robots following the 2D velocity vector $v_i$ output by the proposed controller.

\section{DISCUSSION}
\subsection{Computation Time for Controller Update}
\begin{figure}[b]
    \centering
    \includegraphics[width=\columnwidth]{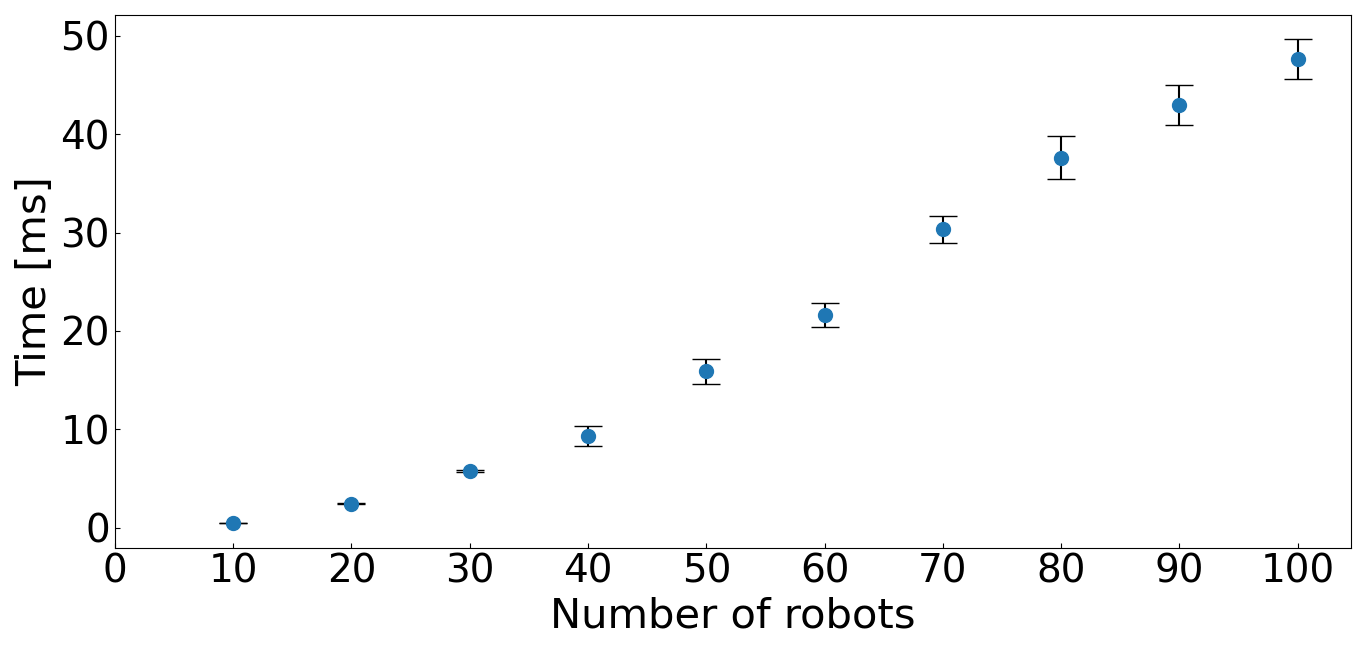}
    \caption{Relationship between the number of robots and the computation time required to update the proposed controller.}
    \label{fig:sph_computation_time}
\end{figure}
To robustly navigate robot swarms using our proposed control method for obstacle-unaware environments, the controller must be updated in real-time when robots detect collisions through indirect collision detection.
Here, we investigate the computation time as a function of the number of robots.

Fig.~\ref{fig:sph_computation_time} shows the processing time per step of the proposed method as a function of the number of robots.
We measured this processing time with a single thread on the PC used in the experiments.
As observed, the processing time for updating the controller for 100 robots using the proposed method is about 50 ms.
This corresponds to 20 Hz in terms of the robot control cycle, indicating that the proposed method can be easily applied to robot swarms comprising 100 robots or less.

In addition, when the number of robots is $N$, the computational complexity of the proposed controller per robot is $O(N)$, as obtained from (\ref{eq:sph_force}), (\ref{eq:force_repulsive}), (\ref{eq:force_position}), and (\ref{eq:force_obstacle}).
Accordingly, the computational complexity for $N$ robots is $O(N^2)$; however, each robot's computational process can be easily parallelized.
For example, if parallelization is implemented with 10 cores, it can be applied to a large swarm system with about 1,000 robots.

\subsection{Application for Pattern Formation}
Here, we evaluated the proposed control method with a focus on the ability of robot swarms to reach a single target point, however, the SPH-based control method can be applied not only to target navigation but also for pattern formation~\cite{Pac2007-op, Pimenta2013-mz}.
This feature is essential in robot swarm applications because they require, for example, the representation of geometric shapes~\cite{Le_Goc2016-fd} and body shapes, such as hands~\cite{Ichihashi2024-gs}, by robot swarm formations.
These applications mainly use RVO-based controllers, which can be immediately adapted to obstacle-unaware environments by replacing them with our proposed controllers.


\section{CONCLUSION}
In this paper, we present a novel control method that significantly improves the navigation of robot swarms in environments with unknown obstacles.
Our approach, which utilizes the SPH model with indirect obstacle detection, enables robot swarms to dynamically adapt to complex scenarios.
We evaluated whether the proposed robot swarm controller allows a swarm to navigate through obstacles without prior obstacle information and the method's efficiency compared with conventional methods. We also demonstrated our method's superior robustness and efficiency compared with conventional methods.
Our results, validated by simulations and real-world experiments, indicate the potential of the proposed method to expand the application range of swarm robotics, especially in areas requiring high adaptability and autonomy.
This research not only advances the field of swarm robotics but also lays the foundation for future research to overcome the unavailability of information about the external environment owing to the lack of sensors.

\bibliographystyle{IEEEtran}
\bibliography{IROS2024_swarm}



\end{document}